# Robust Predictive Modeling Under Unseen Data Distribution Shifts: A Methodological Commentary

**Hanyu Duan,[a] Yi Yang,[a] Ahmed Abbasi,[b,*] Kar Yan Tam[a]**

[a] Department of Information Systems, Business Statistics and Operations Management, School of Business and Management, Hong Kong University of Science and Technology, Clear Water Bay, Kowloon, Hong Kong; [b] Human-centered Analytics Lab and Department of IT, Analytics, and Operations, Mendoza College of Business, University of Notre Dame, Notre Dame, Indiana 46556

*Corresponding author

**Contact:** hduanac@connect.ust.hk, https://orcid.org/0000-0003-2219-7759 (HD); imyiyang@ust.hk, https://orcid.org/0000-0001-8863-112X (YY); aabbasi@nd.edu, https://orcid.org/0000-0001-7698-7794 (AA); kytam@ust.hk, https://orcid.org/0000-0003-3242-0184 (KYT)

**Abstract.** Most research designing novel predictive models, or employing existing ones, assumes that training and testing data are independent and identically distributed. In practice, the data encountered at serving time often deviate from the training distribution, leading to substantial performance degradation and potential design validity and/or biased measurement issues. This challenge is further complicated by the fact that the serving time data are frequently unavailable during model development. This method commentary raises awareness of this overlooked issue through a real-world customer churn example and reviews the growing literature on domain generalization, a subfield of transfer learning that explicitly addresses situations in which the target domain is unseen during training. We further argue for adopting an uncertainty-aware predictive modeling mindset and illustrate how this perspective can be operationalized through the distributionally robust optimization framework. Finally, we offer several practical recommendations to enhance the robustness of predictive modeling under unseen data distribution shifts.

**Funding:** This work was supported by the National Science Foundation [Grant IIS-2039915]. K. Y. Tam was supported by the Research Grants Council of Hong Kong [Grant T35-607/23-N].
**Supplemental Material:** The online appendix is available at https://doi.org/10.1287/isre.2022.0537.

## 1. Introduction

Predictive modeling has become a core methodology for multiple genres of information systems (IS) research, including ones focused on designing predictive artifacts (Shmueli and Koppius 2011, Rai et al. 2017, Agrawal et al. 2018, Padmanabhan et al. 2022) and those using model inferences as inputs in subsequent causal modeling pipelines (Rai 2016, Yang et al. 2018, Padmanabhan et al. 2022). The influential cross-industry standard process for data mining (CRISP-DM) framework has long served as the de facto standard guiding the life cycle of predictive models, progressing from data understanding and preparation to modeling, evaluation, and finally deployment (Chapman et al. 2000). Whereas most academic attention focuses on the earlier stages such as data preparation and model building, it is often at this final stage—when models begin to serve real data—that problems arise. One critical challenge, which forms the central theme of this methodological commentary, is the unseen serving data distribution shift. This phenomenon occurs when the data to which the model actually provides predictions (serving data) differ in distribution from the data used for model training (source data). This challenge has implications for the efficacy of design-related prescriptive knowledge claims (Abbasi et al. 2024) and for biased estimation in explanation-oriented research (Zhang et al. 2023a), opening up research leveraging predictive modeling to the same types of reproducibility and generalizability questions that have surfaced for other genres and methods of inquiry (Open Science Collaboration 2015, Abbasi et al. 2025).

To illustrate, we analyze a real-world customer churn prediction case from a large U.S.-based e-commerce company. Initially, researchers developed a predictive model to identify customers at risk for churn and used it for retention-related decision making. A few months later, when the true churn outcomes became observable, the company found that the model's predictive performance had dropped substantially for new customers acquired after a marketing campaign. A post hoc analysis, as shown in Figure 1, revealed two major types of shifts:

- Covariate shift: The composition of customers changed after the marketing campaign, leading to a distributional difference between the training and serving feature spaces.

- Concept shift: The relationship between certain behavioral features and churn reversed. For example, the number of promotional emails, once positively correlated with churn, became negatively correlated.

The combination of covariate and concept shifts resulted in a substantial decline in predictive accuracy. A detailed quantitative assessment of these distributional shifts is provided in Section 2. In this case, because of continued access to serving data (i.e., visibility into the postdeployment data distribution), the researchers were able to revamp their design and incorporate continual model updates (Kitchens et al. 2018). However, access to serving data beyond the current research test bed is often unavailable (Simester et al. 2020).

In practice, the unseen serving distribution shifts arise from many sources. Firms may launch new marketing campaigns or adopt new targeting channels (Simester et al. 2020), external shocks such as natural disasters can disrupt behavioral patterns (Liu et al. 2023), entering new markets introduces new yet different customer bases (Si et al. 2023), and cross-organizational data inconsistencies frequently occur in enterprise systems (Guo et al. 2022). Compounding this, the serving data are often unavailable or delayed at model training time. For instance, in customer analytics, churn outcomes may only become observable months after prediction (Simester et al. 2020). In enterprise settings, privacy regulations and institutional data silos further restrict access to information from other regions or business units (Van Panhuis et al. 2014, Wilder-James 2016, Hulsen 2020). Table 1 summarizes common real-world predictive tasks and their potential data shift mechanisms for quick reference. When such serving data distribution shifts occur, the performance of predictive models on the serving data often deteriorates. This issue raises important questions for IS researchers, both design researchers who develop novel predictive models for decision support and those who leverage predictive models as part of their causal/explanation-oriented research.

For researchers designing novel predictive models, current modeling and evaluation practices remain largely grounded in the assumption that data are independent and identically distributed (i.i.d.). In model design, the dominant learning paradigm, empirical risk minimization (ERM) (Vapnik 1991), optimizes average predictive accuracy on the training distribution, implicitly assuming that such performance generalizes to i.i.d. test samples. Similarly, model evaluation commonly relies on random data splits or *k*-fold cross-validation, again presuming distributional consistency between training and testing data. These practices overlook the possibility of unseen distribution shifts, leaving model performance under real-world uncertainty largely unexamined. Given the important relation between design-related knowledge claims/abstractions and operational utility, unseen shifts may produce design insights, guidelines, principles, and/or theory that lacks validity (Abbasi et al. 2024).

For researchers using predictive models as tools for constructing machine learning regressors, such as deriving independent or dependent variables from large-scale observational data, distributional shifts can introduce measurement error in model-derived constructs when these models are applied to newly collected data (Yang et al. 2018, Zhang et al. 2023a). Such errors often arise when serving data are collected in periods, platforms, or environments different from those used for model training. This measurement error may bias subsequent econometric analyses, thereby undermining both theoretical validity and empirical robustness of IS research conclusions (Yang et al. 2018; Qiao and Huang 2021, 2025; Zhang et al. 2023a).

**Figure 1.** (Color online) Two Common Types of Data Distribution Shifts

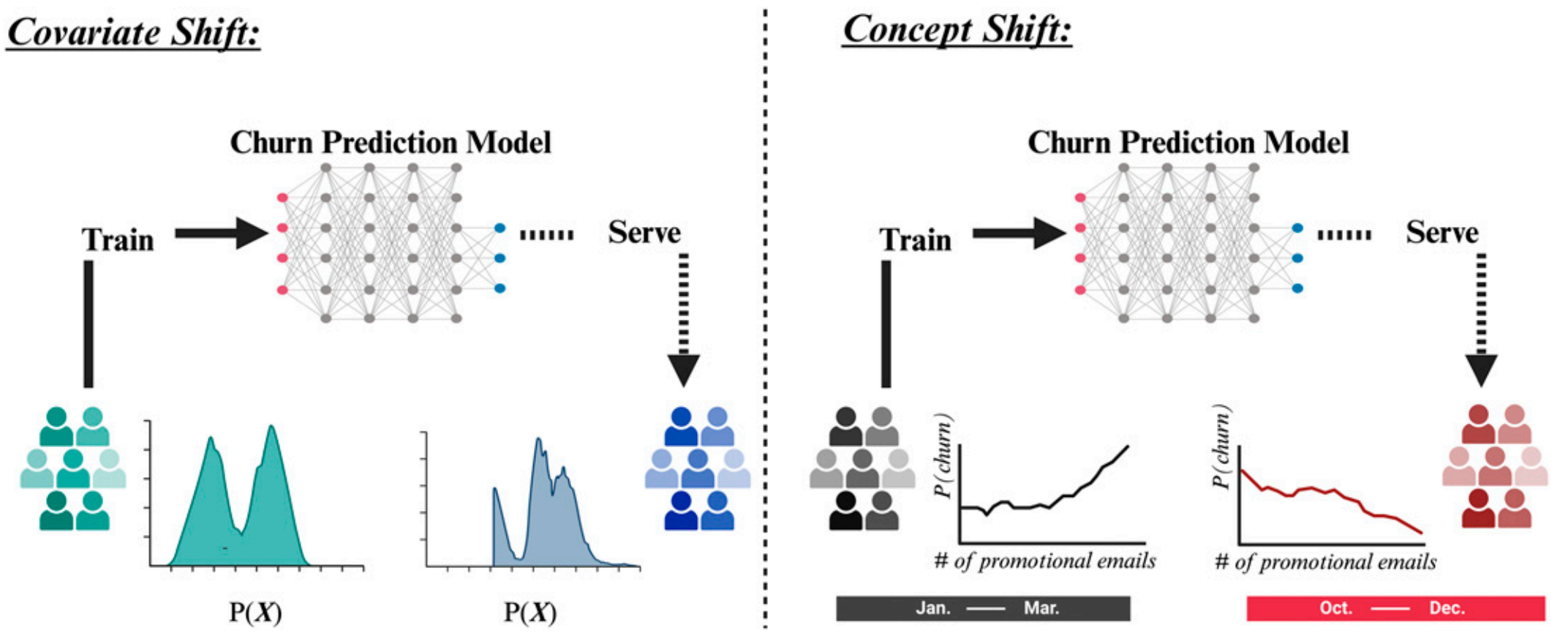

**Table 1.** Examples of Predictive Tasks, Distribution Shift Sources, and Data Availability Challenges

| Predictive task | Source of distribution shift | Why serving data are unavailable at training | Reference |
|---|---|---|---|
| Customer churn and targeting prediction | Rollout of a new marketing campaign changes customer composition and alters churn or conversion behavior | Churn outcomes are observable only months after deployment; serving data collected in a later period | Kitchens et al. (2018) |
| Feature-based pricing in a new market | Entry into a new region changes feature–price relationships and consumer response patterns | Market-level data from the new region are not yet collected at training time | Simester et al. (2020) |
| Cybersecurity threat detection | New attack vectors or system updates create unseen network traffic and behavioral patterns | Zero-day attacks have no prior samples in training data | Ahmad et al. (2023) |
| Hospital readmission prediction | Differences in patient demographics, treatment protocols, and hospital procedures alter readmission likelihood | Developing hospital-specific models using local data are infeasible for many institutions because of privacy and data silo constraints | Hai et al. (2024) |
| Policy learning | Behavioral and contextual shifts between experimental and operational environments affect treatment effectiveness | Only historical experimental data are available; the future serving environment is not observable during model training | Si et al. (2023) |
| Social media sentiment classification | Changes in public opinion alter the semantic association between words and sentiment labels; emergence of new discussion topics changes the vocabulary distribution | Because of the time-series nature of text data, future social media posts are unavailable at training | Guo et al. (2023) |

## 1.1. Purpose of This Method Commentary

This methodological commentary serves four key purposes.

First, it seeks to shed light on the problem of serving data distribution shift. We clarify its defining characteristics, including covariate shift, concept shift, and their combinations, and discuss its relevance for IS research. Specifically, we use a real-world customer churn prediction task to illustrate how a marketing campaign rollout can alter customer behavior and induce both covariate and concept shifts, ultimately degrading model performance on the serving data. In Online Appendix A, we also note how this problem can manifest in unstructured data settings that are becoming increasingly prevalent for research employing predictive models (Ahmad et al. 2020, Yang et al. 2023).

Second, we aim to situate existing methodological approaches for tackling the serving distribution shift challenge, particularly domain generalization (DG), a subfield of transfer learning. Domain generalization aims to train models that can effectively generalize to unseen domains by leveraging knowledge from multiple training domains.[1] We review this line of work and provide an accessible taxonomy-based categorization of DG methods. In this regard, this work complements Simester et al. (2020), which empirically documents the challenges of predictive modeling under unseen serving data shifts but does not review methodological solutions. In this methodological commentary, we extend their insights by offering a structured overview of DG methods, aiming to help IS researchers and practitioners better understand, compare, and potentially advance these approaches for robust predictive modeling.

Third, we advocate a shift in mindset from average performance–driven predictive modeling to uncertainty-aware predictive modeling. Rather than solely optimizing for average accuracy on training data, as grounded in empirical risk minimization, we conceptually and empirically underscore the importance of considering reliable performance under plausible but unseen distributional shifts. We further demonstrate how this uncertainty-aware mindset can be implemented through a distributionally robust optimization (DRO) framework, which focuses on minimizing worst case loss rather than maximizing average performance on training distributions. Following Rai (2020), proactively managing uncertainty provides a pathway for systematically advancing IS research. Such an uncertainty-aware mindset not only motivates design researchers to develop more robust and practically relevant predictive artifacts but also encourages empirical researchers to mitigate potential measurement errors and estimation biases in theory development.

Finally, we offer practical recommendations for researchers on how to improve robustness throughout the predictive modeling life cycle by following a three-step approach of assessing, applying, and evaluating. This approach includes identifying potential data shifts early in study design, selecting appropriate DG methods to safeguard uncertainty, simulating data sets that reflect realistic shifts, conducting stress tests under these simulated shifts, and reporting both average and worst case results. By embedding risk awareness throughout the entire modeling process, researchers can enhance the robustness and practical relevance of research involving predictive models in real-world enterprises and socioeconomic contexts.

## 2. A Real-World Customer Churn Prediction Problem

We illustrate serving data distribution shift and its impact on predictive modeling through a real-world customer relationship management context. Customer analytics is considered an important problem space for IS research (Chen et al. 2012). The setting involves a large U.S.-based e-commerce company whose primary predictive analytics objective is to predict customer churn: whether a customer will stop purchasing within one year of the customer's initial transaction. This predictive task aims to improve customer retention and optimize marketing expenditures. The data set is collected from the company's customer analytics database.[2] It contains detailed behavioral and demographic data for customers who made their first purchase, covering 368 features across seven categories: demographics, transactions, choice, messaging, channel, engagement, and satisfaction. Following Kitchens et al. (2018), we construct input features based on customer activity during the first 30 days after the initial purchase and define the target variable as a binary churn indicator observed within the subsequent 365 days. A customer is labeled as churn if no repeat purchase occurs during that period and as not churn otherwise.

The serving distribution shift in this study arises naturally from a real marketing intervention that occurred during the observation period. In September 2013, the company developed a customer churn prediction model using data that included customers whose initial purchases occurred between January and September 2012. Because churn outcomes are observed one year after the initial purchase, customers who purchased between October 2012 and September 2013 did not yet have observable churn labels and were, therefore, excluded from model training. The trained model was subsequently deployed to predict churn for customers who made their first purchases between October 2013 and February 2014. Notably, in October 2012, outside the period covered by the training data, the firm launched a major "buy one, get two free" promotion, its first campaign of this kind. Prior to this change, all promotional offers involved discounts of 20% or less. Therefore, in this setting, the training data consist of customers who made their purchases before the campaign rollout, whereas the serving data comprise customers who made their purchases after the campaign.

### 2.1. Evidence of Serving Data Distribution Shift

The shift in marketing strategy created a natural serving data distribution shift: customers acquired after the campaign differed from those in the training data, leading to a covariate shift in feature distributions, whereas the relationship between customer characteristics and churn outcomes also evolved, reflecting a concept shift.

**2.1.1. Covariate Shift.** This shift occurs when the distribution of input features differs between the training and serving environments. Here, we examine how the distribution of customer features changed following the rollout of the marketing campaign. To visualize these changes, we employ t-distributed stochastic neighbor embedding (t-SNE) (Van der Maaten and Hinton 2008) to project both the training (precampaign) and target (postcampaign) data sets into a two-dimensional feature space. The resulting visualization, shown in Figure 2, reveals a clear separation between the two data sets: observations from the prerollout period (circle) and postrollout period (triangle) occupy distinct regions of the latent space with limited overlap. This divergence indicates that the marginal distribution of input features shifted between the two periods.

**Figure 2.** (Color online) t-SNE Visualization of Customer Feature Distributions

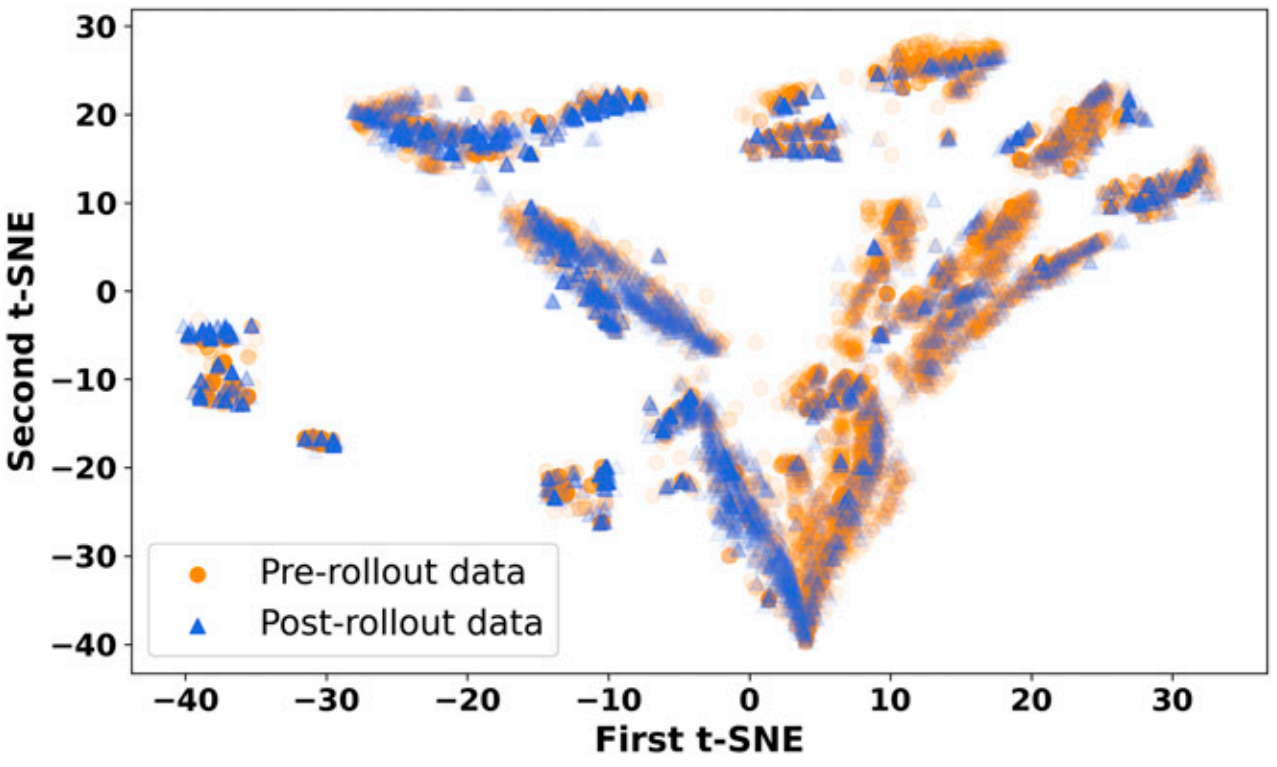

**2.1.2. Concept Shift.** This shift occurs when the conditional relationship between inputs and outcome target variable changes. Here, we apply interpretable machine learning techniques to quantify the potential concept shift, that is, how the relationship between input features and customer churn evolves over time. In particular, we compute Shapley additive explanation (SHAP) values (Lundberg and Lee 2017) for each feature using two churn prediction models: one trained on the source data (January–September 2012) and another trained on the serving data (September 2013–February 2014). Figure 3 presents the 10 most influential features along with their SHAP value distributions. The left panel displays the results for the model trained on source data and the right panel for the model trained on serving data. As highlighted by the black outlines, the feature `maxOfferFullPriceDiscount_email` exhibits a clear reversal in its association with churn. In the training period, higher values of this feature, reflecting customers who received large full-price discounts via email, are associated with an increased likelihood of churn, suggesting that aggressive discounts might have signaled disengagement. In contrast, in the serving period, after the rollout of the marketing campaign, this relationship reverses: the same feature becomes negatively associated with churn, implying that such discounts began to serve a retention rather than a disengagement role. Several other features also displayed changes in both the magnitude and direction of their influence as indicated by the variation in SHAP value distributions across the two models.

## 2.2. Empirical Evidence of Performance Degradation

We now examine how such serving data distribution shift impacts the customer churn prediction performance. We evaluate both a neural network and an XGBoost classifier (Chen and Guestrin 2016), both trained via classical empirical risk minimization. The performance results are reported in Table 2. As evidenced by the significant gap between prerollout and postrollout performance (approximately a 7% performance drop), the model trained with prerollout data struggles to generalize well to postrollout data, primarily because of the shift in customer behavior caused by the campaign rollout. Given that such prediction performance metrics are used to validate contributions when designing predictive artifacts (Abbasi et al. 2024) or when developing machine learning–based variables/regressors (Yang et al. 2018, Zhang et al. 2023a), the methodological implications of unseen distribution shifts for scientific validity are extremely consequential. It is imperative to develop robust predictive models that can mitigate the impact of potential unseen serving data distribution shifts. Accordingly, in the remainder of this methodological note, we present an accessible taxonomy of the DG literature, underscore the importance of DG for uncertainty-aware predictive modeling, illustrate how DRO—an important type of DG—offers a pathway for such modeling, and present practical recommendations for IS researchers.

# 3. Domain Generalization: A Bird's-Eye View of the Landscape

Predictive models are valuable only to the extent that they generalize beyond the data used to train them. Whereas classic i.i.d. train–test splits provide a convenient proxy, they may not reflect deployment realities. This motivates substantial interest in transfer learning (Pan and Yang 2009), whose core intuition, tracing back to early work on generalization (Judd 1932), is that learning should extract abstractions that transfer across situations rather than memorize patterns.

DG is a branch of transfer learning that trains models to perform well on previously unseen deployment environments (Shen et al. 2021, Wang et al. 2022). Unlike domain adaptation or semisupervised learning,

**Figure 3.** (Color online) SHAP Value Distributions of Top Features

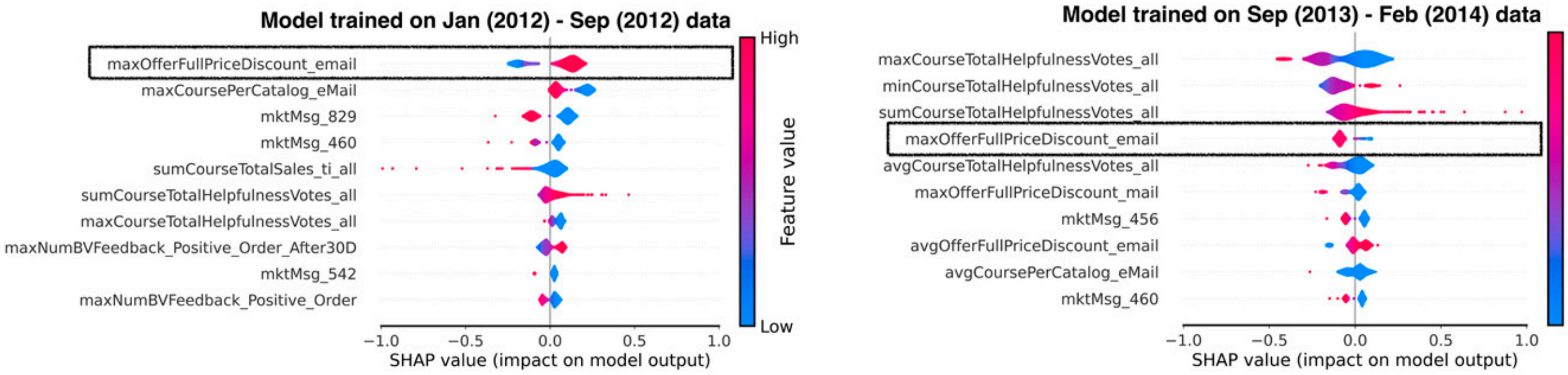

*Note.* Churn prediction models trained on source data (left) and serving data (right); the 10 most influential features, ranked by SHAP importance, are displayed from top to bottom.

**Table 2.** Customer Churn Prediction Performance

| | Classification model | |
|---|---|---|
| Evaluation stage | Neural network | XGBoost |
| Prerollout phase | 0.729 | 0.736 |
| Postrollout phase | 0.676 (7% ↓) | 0.686 (7% ↓) |

*Note.* Performance is measured by AUROC.

DG assumes no access to target-domain data during training (not even unlabeled inputs) (Glorot et al. 2011). This makes DG especially relevant in early deployment stages when the future serving distribution is unknown.

**Definition 1** (Domain). Let $\mathcal{X}$ denote the feature space and $\mathcal{Y}$ the label space. A domain is a joint distribution $P_{\mathbf{X}Y}$ over $(\mathbf{X}, Y)$ with $\mathbf{X} \in \mathcal{X}$ and $Y \in \mathcal{Y}$.

**Definition 2** (Domain Generalization). Given a training data set consisting of $K$ source domains $\mathcal{S} = \{\mathcal{S}^i\}_{i=1}^K$, where $\mathcal{S}^i = \{(\mathbf{x}_j^i, y_j^i)\}_{j=1}^{n_i}$ and $P_{\mathbf{X}Y}^i \neq P_{\mathbf{X}Y}^j$ for $i \neq j$, DG aims to learn a function $f_\theta : \mathcal{X} \to \mathcal{Y}$ from $\mathcal{S}$ that maximizes predictive performance on an unseen target domain $\mathcal{T} \sim P_{\mathbf{X}Y}^T$ with $P_{\mathbf{X}Y}^T \neq P_{\mathbf{X}Y}^i$ for all $i \in \{1, \ldots, K\}$. The target domain is inaccessible during training.

Following the convention in the domain generalization literature (Wang et al. 2022, Zhou et al. 2022, Khoee et al. 2024), we assume that the training data are composed of multiple source domains ($K > 1$), each representing a distinct data-generating process (e.g., region, time, or organization). Although the single-domain case ($K = 1$) is possible, it has received limited attention in prior DG studies because it is less common in real-world settings (Qiao et al. 2020). Therefore, we adopt the standard multidomain definition to maintain consistency with the existing DG literature. In practice, the definition of domains is task-dependent, and the number of source domains $K$ can be either naturally given or constructed from the data. For example, when data come from multiple regions, each region naturally forms a domain with its own data-generating process. When data arise from a single organization over time (e.g., customer churn data collected before a policy change), the training set may be partitioned into several time-specific domains to approximate such distributional heterogeneity. In both cases, the goal is to learn from the available source domains (training) and generalize to an unseen target domain that reflects a new environment (serving).

## 3.1. A Multilevel Taxonomy of DG Methods

A useful way to organize the DG landscape is by where in the learning pipeline a method intervenes: the input level (data), the representation level (features), or the learning-process level as shown in Figure 4. Figure 5 conceptually illustrates how different classes of DG methods achieve generalization.

**Figure 4.** Conceptual Organization of Domain Generalization Methods

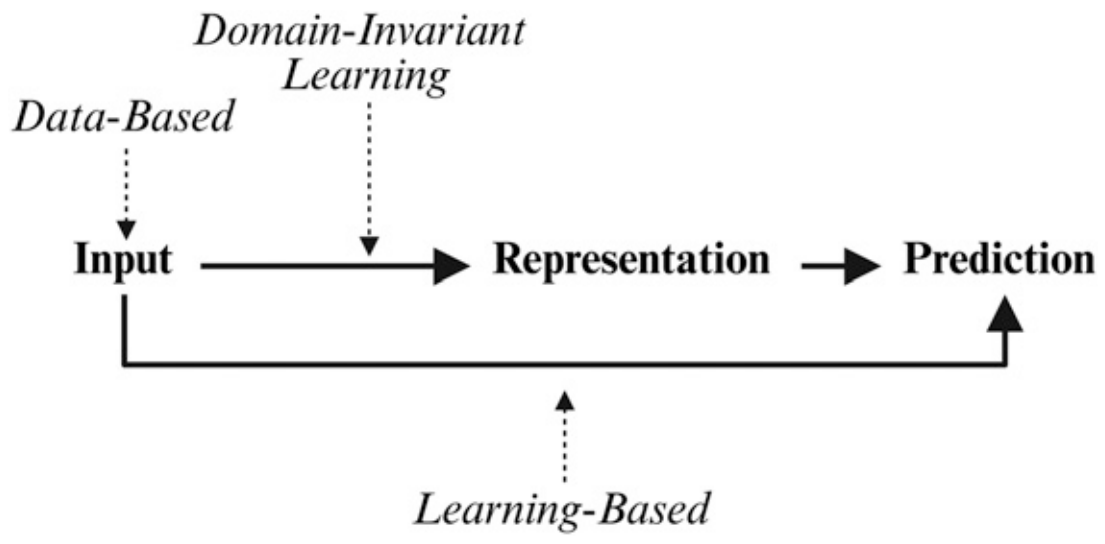

*Note.* Methods are organized according to where they intervene in the learning pipeline.

### 3.1.1. Input-Level: Data-Based Approaches.

Data-based approaches aim to enrich the diversity of training samples by generating synthetic data or performing data manipulation and augmentation. Most of these methods originate from the computer vision literature, in which image data are naturally amenable to controlled perturbations. Typical strategies include data augmentation, which perturbs the input space through transformations such as noise injection, feature masking, or synthetic resampling, and style or feature manipulation, which alters statistical properties of the data (e.g., color, texture, or feature covariance) to simulate domain shifts (Carlucci et al. 2019, Xu et al. 2021, Zhou et al. 2021).

Additionally, generative models such as variational autoencoders and generative adversarial networks have been employed to synthesize artificial training examples with varied styles or backgrounds, thereby increasing domain diversity (Rahman et al. 2019, Qiao et al. 2020, Li et al. 2021). More recently, domain interpolation techniques such as Mixup (Zhang et al. 2018) and its variants have gained attention. These methods create new samples by linearly combining features or representations from different domains, effectively smoothing the decision boundary across heterogeneous training distributions. The central intuition behind these approaches is to expose the model to a broader range of possible variations during training, encouraging it to learn domain-agnostic, shareable features that generalize beyond the observed source domains. Using CRISP-DM speak, these methods operate at the data understanding and preparation stages.

### 3.1.2. Representation-Level: Domain-Invariant Learning.

Another major family of domain generalization methods operates at the representation level, aiming to learn feature spaces that capture the stable, causal structure of a task rather than superficial domain-specific artifacts. The guiding principle is that, if representations

**Figure 5.** (Color online) Illustration of Generalization Mechanisms in Domain Generalization Methods

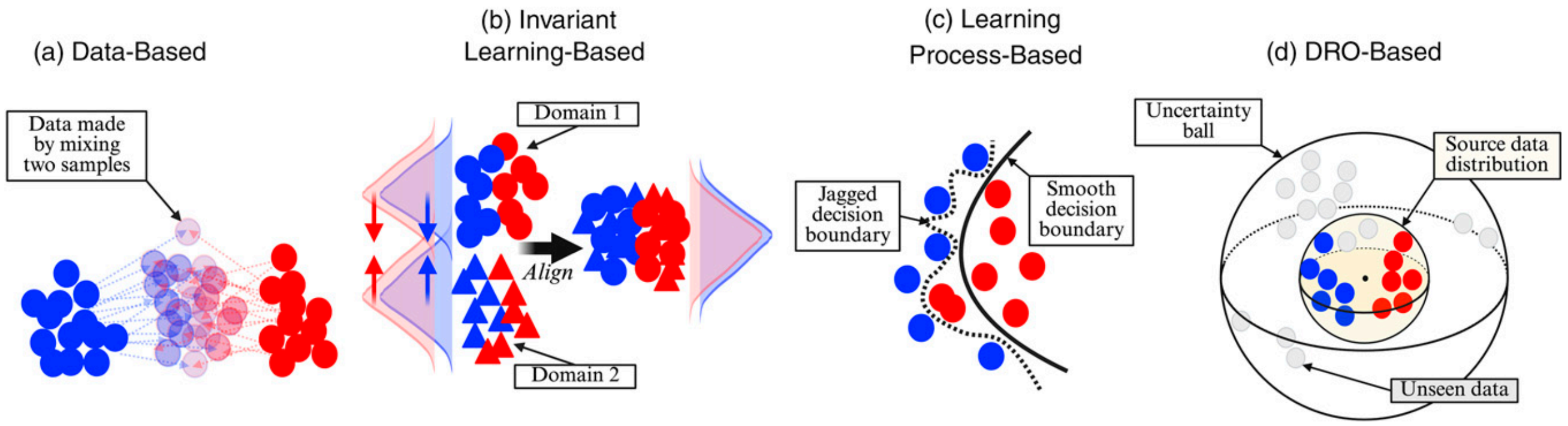

*Notes.* (a) Synthetic data are generated to increase training data diversity, (b) domain-invariant features are learned via representation alignment, (c) certain training strategies (e.g., regularization) are applied to learn smooth decision boundaries and improve robustness, and (d) the optimization procedure incorporates an additional uncertainty region to improve robustness to previously unseen data distributions. Each marker represents a data point, with shape (circle versus triangle) indicating domain membership.

encode invariant mechanisms across environments, a single predictive model can generalize reliably to unseen domains.

Some work in this stream focuses on representation alignment, ensuring that feature distributions from different domains become indistinguishable. Examples include adversarial alignment methods such as domain adversarial neural networks (Ganin et al. 2016) and moment-based alignment approaches such as correlation alignment (Sun and Saenko 2016), which minimize statistical discrepancies between domains. More principled formulations, such as invariant risk minimization (IRM) (Arjovsky et al. 2019), directly enforce that the optimal predictor remains consistent across environments, thereby learning features predictive of the outcome in all domains. Beyond alignment, several approaches seek to uncover more fundamental invariances. Causal representation learning methods (Mahajan et al. 2021, Sheth et al. 2022) attempt to identify stable causal mechanisms that persist under interventions, offering stronger guarantees of transferability than purely statistical alignment.

Additionally, a complementary direction uses representation disentanglement to explicitly separate domain-invariant and domain-specific factors. Models such as DIVA (Ilse et al. 2020) and MD-Net (Wang et al. 2020) employ variational autoencoders or generative adversarial networks to isolate invariant latent variables, suppressing nuisance variation. Similarly, normalization-based methods (e.g., Seo et al. 2020) remove domain-specific statistical biases through adaptive feature standardization.

#### 3.1.3. Learning Process Level: Learning-Based Approaches.

A third major class of domain generalization methods operates at the learning process level, focusing on how models are trained rather than on the data or feature representations themselves. The core idea is to design training strategies that explicitly encourage robustness to unseen domains by simulating, regularizing, or averaging over potential distributional shifts during learning. As in the representation level, learning process level approaches happen at the modeling stage of CRISP-DM.

A prominent line of work adopts meta-learning, in which models are trained to anticipate domain shifts through episodic learning. For example, meta-learning domain generalization (MLDG) (Li et al. 2018) constructs virtual train–test splits among source domains, updating model parameters on one subset to ensure performance on another, thereby simulating exposure to new domains. MetaReg (Balaji et al. 2018), by contrast, learns a meta-regularizer that shapes the optimization landscape to favor generalizable solutions. These methods effectively teach the model how to learn from domain variation itself. Beyond meta-learning, a variety of regularization-based approaches constrain the optimization process to prevent overfitting to domain-specific artifacts (Cha et al. 2021, Gulrajani and Lopez-Paz 2021).

#### 3.1.4. Limitations of Existing Approaches.

Whereas all of the above methods have been evaluated under domain generalization settings, it is important to recognize their limitations and consider their practical applicability in real-world research environments.

i. Predominant focus on covariate shift: Most DG methods are designed to learn domain-invariant features that generalize across training and unseen domains as exemplified by IRM and causal representation learning. A canonical illustration is the cow-on-grass problem: learning that the invariant concept is the cow itself, so the model can correctly recognize a cow on snow or ice. This design objective is well-suited for handling covariate shift, in which only the input distribution $P(\mathbf{X})$ changes, whereas $P(Y|\mathbf{X})$ remains stable.

However, in real-world managerial contexts, serving data distribution shifts rarely occur in isolation. They typically involve the combination of both covariate and concept shifts. This combined form of shift could limit the effectiveness of existing DG methods. Current benchmarking environments such as DomainBed (Gulrajani and Lopez-Paz 2021) and TableShift (Gardner et al. 2024) further reinforce this bias as they are intentionally designed so that domain-invariant features exist and can be learned. As a result, these benchmarks primarily measure performance under covariate shift, leaving robustness under concept shift largely untested (Zhang et al. 2023b).

ii. Average-performance mindset: A second limitation lies in the underlying mindset of optimizing for average performance. Although most DG studies explicitly acknowledge the existence of serving distribution shifts, their training objectives rarely incorporate a notion of uncertainty. From a managerial and operational standpoint, predictive modeling under uncertainty is a problem of research robustness and risk management. Addressing this challenge requires not only new DG techniques but also a shift in mindset from optimizing for average performance to safeguarding performance under uncertain conditions. One can draw an analogy with portfolio management: portfolio managers do not evaluate investments solely based on expected returns but also on potential losses (such as maximum drawdown) under unfavorable market conditions. Similarly, researchers employing predictive modeling under potential serving data distribution shifts should consider incorporating uncertainty considerations into both model training and evaluation, drawing inspiration from the rich scientific traditions of design validity (rigor when prescribing) and robustness checks (rigor when explaining).

### 3.2. Distributionally Robust Optimization: Optimizing for Uncertainty

One stream of the learning-based approaches incorporates uncertainty at the training objective level via DRO (Ben-Tal et al. 2013, Rahimian and Mehrotra 2019, Duchi et al. 2021). Instead of minimizing expected loss (average performance) on the training distribution as in the classical ERM framework, DRO explicitly accounts for possible distributional shifts by optimizing performance under the worst case scenario within a neighborhood of the training data distribution:

$$\underset{\theta\in\Theta}{minimize} \sup_{H\in\mathcal{H}} \mathbb{E}_H[\ell(\theta;(\mathbf{X},Y))]. \tag{1}$$

Here, $\theta \in \Theta$ represents the learned parameters of the predictive model, and $\ell : \mathcal{X} \times \mathcal{Y} \to \mathbb{R}$ denotes the loss function of a specific prediction task, such as cross-entropy loss for classification or mean squared error for regression tasks. The set $(\mathcal{H})$ is a hypothetical uncertainty set representing a collection of plausible distributions.

DRO provides a promising framework for incorporating uncertainty into the domain generalization problem. This formulation offers two key advantages. First, the uncertainty set in DRO provides a flexible way to model both covariate and concept shifts. Second, it represents a mindset shift from optimizing for average performance to optimizing for the worst case scenario. Several domain generalization methods have built upon the DRO framework by defining different forms of the uncertainty set that characterizes potential distributional shifts (Volpi et al. 2018, Qiao et al. 2020, Sagawa et al. 2020). The quality of a predictive model trained under the DRO framework depends on how the uncertainty set is specified. If the uncertainty set is too broad, the resulting model becomes overly conservative and may underfit the data; if it is too narrow, it may fail to capture meaningful distributional variations. Consequently, designing practically meaningful uncertainty sets remains an open methodological challenge for real-world predictive modeling tasks.

## 4. Illustrating Robust Predictive Modeling Under Distributional Shifts

In this section, we illustrate how the distributionally robust optimization framework can operationalize the uncertainty-aware mindset discussed above. Following prior commentaries (Abbasi et al. 2025), rather than proposing a new method, our goal is to demonstrate how uncertainty can be implemented in predictive modeling to account for complex real-world shifts, particularly those that combine both covariate and concept components.

### 4.1. Operationalizing DRO for Covariate and Concept Shifts

**4.1.1. Define the Uncertainty Set.** To capture both covariate and concept shifts, we specify, for each source domain $\mathcal{S}^i, i \in \{1, \ldots, K\}$, a domain-level uncertainty set:

$$\mathcal{H}^i = \{H^i : C_{\text{cov}}(\mathcal{S}^i, H^i) \leq \rho_1, C_{\text{conc}}(\mathcal{S}^i, H^i) \leq \rho_2\}, \tag{2}$$

where $C_{\text{cov}}(\cdot,\cdot)$ measures covariate shift between the source distribution and an uncertainty distribution, $C_{\text{conc}}(\cdot,\cdot)$ measures concept shift, and $\rho_1, \rho_2 > 0$ are the radius parameters controlling the size of the uncertainty region. Larger values indicate broader space and, thus, potentially more adverse worst case scenarios.

For each source domain, the DRO objective (Equation (1)) can be written via a Lagrangian relaxation as

$$\min_{\theta^i\in\Theta} \sup_{H^i\in\mathcal{H}^i} \{\mathbb{E}_{(\mathbf{X},Y)\sim H^i}[\ell(\theta^i;(\mathbf{X},Y))] - \gamma_1 C_{\text{cov}}(\mathcal{S}^i, H^i) - \gamma_2 C_{\text{conc}}(\mathcal{S}^i, H^i)\}, \tag{3}$$

where $\theta^i$ parameterizes the predictive model trained on domain $\mathcal{S}^i$ and $\gamma_1, \gamma_2 \geq 0$ are penalty weights for the covariate and concept shift terms, controlling the degree of uncertainty set shifts, respectively. Next, we unpack the components inside the inner maximization term to clarify the role of each in characterizing the worst case distribution.

**4.1.2. Adversarial Term $\ell(\boldsymbol{\theta}^i; (\mathbf{X}, \boldsymbol{Y}))$.** This term measures the loss of individual samples under model $\theta^i$. During the inner maximization, this term drives the optimization process to identify samples that maximize the model's prediction error, typically those lying near the current decision boundary of $\theta^i$. This is analogous to adversarial training, in which the model is optimized to be robust against adversarial perturbations (Goodfellow et al. 2014, Madry et al. 2018, Sinha et al. 2018).

**4.1.3. Covariate Shift Term $C_{\mathrm{cov}}$.** Covariate shift refers to changes in the input distribution $P(\mathbf{X})$, keeping the conditional relationship $P(Y|\mathbf{X})$ fixed. Following Volpi et al. (2018), we define the distance between the uncertainty distribution $H^i$ (characterized by covariate shift) and the source domain $\mathcal{S}^i$ as

$$C_{\mathrm{cov}}(\mathcal{S}^i, H^i) = \inf_{\Pi(\mathcal{S}^i, H^i)} \mathbb{E}_{(\mathbf{x},y),(\mathbf{x}^*,y^*)\sim\Pi}\left[\frac{1}{2}\|\theta^i(\mathbf{x}) - \theta^i(\mathbf{x}^*)\|_2^2 + \infty \cdot \mathbf{1}\{y \neq y^*\}\right], \tag{4}$$

where $\Pi(\mathcal{S}^i, H^i)$ denotes all couplings between the two distributions and $\theta^i(\mathbf{x})$, $\theta^i(\mathbf{x}^*)$ are the latent representations of $\mathbf{x}$ and $\mathbf{x}^*$, respectively. Here, the term $\infty \cdot \mathbf{1}\{y \neq y^*\}$ enforces label consistency, allowing feature perturbations that preserve representation similarity. This term indicates that the uncertainty set $\mathcal{H}^i$ covers distributions that share feature perturbations with the source domain $\mathcal{S}^i$, preserving label consistency. During the inner maximization, this term drives the optimization process to explore feature perturbations that shift the input distribution, maintaining label consistency. Intuitively, it encourages the generation of samples in the neighborhood of the original data, effectively expanding the convex hull of the training distribution, to ensure that the learned decision boundary remains stable under small but plausible feature variations (i.e., covariate shift).

**4.1.4. Concept Shift Term $C_{\mathrm{conc}}$.** Concept shift occurs when the mapping $P(Y|\mathbf{X})$ changes, whereas $P(\mathbf{X})$ remains relatively stable. Prior DRO-based DG methods have not explicitly modeled concept shift (Volpi et al. 2018, Sagawa et al. 2020). Here, we propose to quantify the degree of concept shift between two domains by measuring the predictive loss of a model trained on one domain when evaluated on another. Specifically, the uncertainty distribution characterized by concept shift with respect to source domain $\mathcal{S}^i$ is defined through the loss incurred by an auxiliary model $\theta^j$ trained on a different domain $\mathcal{S}^j$ $(j \neq i)$:

$$C_{\mathrm{conc}}(\mathcal{S}^i, H^i) = \inf_{\Pi(\mathcal{S}^i, H^i)} \mathbb{E}_{(\mathbf{x},y),(\mathbf{x}^*,y^*)\sim\Pi}[\ell(\theta^j; (\mathbf{x}^*, y^*))], \tag{5}$$

where $\theta^j$ is an auxiliary model trained on another source domain $\mathcal{S}^j$ $(j \neq i)$. This term indicates that the uncertainty set $\mathcal{H}^i$ covers distributions that deviate from the source domain $\mathcal{S}^i$ and approach regions near the decision boundary of another domain $\mathcal{S}^j$. During the inner maximization, this term drives the optimization process to identify samples that incur low loss under the auxiliary model $\theta^j$, revealing areas in which different domains disagree in their label-generation mechanisms. Intuitively, encouraging robustness in such disagreement regions helps the model learn a decision boundary that is less sensitive to domain-specific labeling rules, effectively smoothing the boundary across domains and preventing overfitting to any single source domain.

**4.1.5. Discussion.** It is worth noting that our operationalization of DRO departs from the conventional practice of defining uncertainty sets as abstract divergence balls, such as Wasserstein or Kullback–Leibler balls around the empirical training distribution (Namkoong and Duchi 2016, Sinha et al. 2018, Gao and Kleywegt 2023). Instead, we explicitly construct the uncertainty set to capture distributional variations that are expected in the focal predictive task, namely, covariate and concept shifts. Whereas classical ball-based formulations provide elegant generalization guarantees, our formulation complements them by explicitly aligning the uncertainty set with domain-relevant sources of variation. Consistent with the view of Rahimian and Mehrotra (2019), who emphasize that the shape of the uncertainty set should reflect the modeler's perception of uncertainty and problem structure, our goal is not to establish new theoretical bounds but to illustrate how the uncertainty-oriented optimization principle of DRO can be operationalized in a task-specific manner.

**4.1.6. Optimization Procedure.** The objective in Equation (3) can be reformulated as a two-step min–max problem. In the inner maximization step, each sample is perturbed within the uncertainty region defined by the covariate and concept shift constraints, effectively performing a data-augmentation process. This step generates a set of fictitious samples that represent worst case scenarios: inputs that induce high loss under the current model parameters. In the outer minimization step, the model parameters are updated via stochastic gradient descent or other optimization approaches to minimize the expected loss over both the original training samples and these worst case augmented samples. Although this optimization procedure structurally resembles ERM, it differs in that the learning process is carried out on an enriched training distribution: one that explicitly explores low-density, high-loss regions near the decision boundary.

### 4.2. Hyperparameter Selection via Leave-One-Domain-Out Cross-Validation

In the proposed DRO framework, the hyperparameters $\gamma_1$ and $\gamma_2$ control the relative strength of the covariate shift and concept shift components within the uncertainty set. Conceptually, they are inversely related to the radius parameters $\rho_1$ and $\rho_2$ that define the extent to which the uncertainty distribution $H^i$ can deviate from the source domain $\mathcal{S}^i$ (Equation (2)). A larger $\gamma$ value corresponds to a smaller uncertainty set, whereas a smaller $\gamma$ allows for greater deviation. Selecting these parameters entails a fundamental trade-off between robustness and performance. If the uncertainty set is too large (i.e., $\gamma$ too small), the model may overemphasize worst case scenarios, degrading average-case performance when the deployment shift is mild. Conversely, if the uncertainty set is too small (i.e., $\gamma$ too large), the model may fail to safeguard against meaningful shifts. This trade-off resembles paying an insurance premium (Si et al. 2023), in which one sacrifices a small amount of performance under normal conditions to gain protection when adverse events occur.

Given that the target domain is unavailable during training, it is difficult to determine an appropriate level of robustness a priori. Following the practice in the DRO literature (Rahimian and Mehrotra 2019), we adopt a data-driven, leave-one-domain-out cross-validation strategy (Gulrajani and Lopez-Paz 2021) to tune $\gamma_1$ and $\gamma_2$. Specifically, given $K$ source domains, we iteratively hold out one domain $\mathcal{S}^k$ for validation and train the model on the remaining $K-1$ domains. For each iteration, $\gamma_1$ and $\gamma_2$ are selected to minimize the DRO objective, maintaining good predictive performance on the held-out domain $\mathcal{S}^k$. This process is repeated across all domains, and the final hyperparameters are determined by averaging the validation performance across all folds.

Alternative approaches can also be considered. For instance, Volpi et al. (2018) propose an ensemble-based heuristic that trains multiple models with different $\gamma$ values, each corresponding to a different level of robustness. At inference time, the model with the highest softmax confidence for a given input is selected. Overall, this cross-validation based tuning or heuristic selection serves as a practical approach for researchers to balance robustness and generalization in distributionally robust predictive modeling, particularly when the serving environment is unknown at training time.

### 4.3. Empirical Evaluation of DRO and DG Methods for Real-World Predictive Tasks

In this section, we show the performance of the DRO and other DG methods under complex real-world distribution shifts using two predictive tasks. The first task focuses on customer churn prediction in a marketing context (as discussed in Section 2), and the second examines in-hospital mortality prediction in healthcare settings. Both data sets used in this study are structured tabular data. In Online Appendix A, we further examine the performance of DG methods on an unstructured text-based prediction task.

**4.3.1. Test Beds.** (i) Customer churn data set: The first evaluation revisits the customer churn prediction task introduced earlier in Section 2. The training data (before September 2013) are drawn from precampaign customer cohorts, whereas the serving data (September 2013–February 2014) correspond to the postcampaign period. The marketing campaign causes both covariate and concept shifts as it alters customer composition and churn behavior. (ii) Hospital intensive care unit (ICU) mortality data set: The second evaluation is based on the multicenter eICU Collaborative Research Database (Pollard et al. 2018), which contains ICU records from hospitals in four regions of the United States: Midwest, South, West, and Northeast. The task is to predict whether a patient will die in the hospital given clinical data collected within the first 24 hours of the ICU stay. Each region represents a separate domain that differs in patient demographics, clinical protocols, and equipment configurations, resulting in both covariate and concept shifts (Guo et al. 2022). In each experiment, three regions are used as source domains for training and one as the unseen target domain for testing. The data set includes approximately 8,000 patient cases with 10 temporal and three static clinical features, and about 10% of patients experience in-hospital mortality, indicating a moderately imbalanced outcome distribution. This setup reflects a realistic domain generalization scenario in which a predictive model trained in certain hospitals needs to generalize to new hospitals with different patient populations and medical practices.

**4.3.2. Predictive Models and Evaluation Metrics.** For the customer churn prediction model, we use a multilayer perceptron neural network (MLP) as the base predictive model. The network comprises an input layer with 368 dimensions, four hidden layers containing 64, 32, 16, and 8 neurons, respectively, and an output layer with two dimensions. The hidden layers use the rectified linear unit activation function (Agarap 2018). For mortality prediction, because each patient record is a time series, we employ a long short-term memory (LSTM) network (Hochreiter and Schmidhuber 1997). LSTM is suitable for sequential data modeling, and it has been used to effectively model time series data for healthcare outcome predictions, including mortality prediction (Choi et al. 2016, Rajkomar et al. 2018). The LSTM model has 128 hidden units, and the output vector of the last hidden unit is fed into a linear layer for final outcome prediction. We concatenate the temporal features with the static features as the input for each LSTM time step and use 20 time steps (i.e., sequence length). Because both prediction tasks are binary classification problems, we use the area under the receiver operating characteristic curve (AUROC) as the evaluation metric. The AUROC value ranges from zero to one, in which higher values indicate better overall predictive performance.

**4.3.3. ERM and DG Methods.** We evaluate the following predictive modeling methods. All methods are trained on the same set of $K$ source domains using the same predictive architecture (either MLP or LSTM as defined earlier), ensuring that they have identical information and no access to serving-domain data during training.

i. ERM (Vapnik 1991) minimizes the average empirical loss over all source domains without explicitly accounting for domain shifts. Although it does not incorporate robustness considerations, ERM remains the most widely used approach in predictive modeling and often achieves competitive results in practice (Cha et al. 2021, Zhang et al. 2021b, Guo et al. 2022, Pfohl et al. 2022).

ii. We consider two data-based DG approaches. Adversarial data augmentation (ADA) (Volpi et al. 2018) improves model resilience by generating adversarially perturbed samples that mimic distributional shifts. Mixup (Zhang et al. 2018) linearly interpolates pairs of training samples and their labels to create smoothed synthetic data, which helps regularize the model and mitigate overfitting to specific domains.

iii. IRM (Arjovsky et al. 2019) encourages the model to learn representations that yield consistent predictive relationships across domains, targeting the stable mechanisms underlying $P(Y|\mathbf{X})$.

iv. We consider three learning-based DG approaches. MetaReg (Balaji et al. 2018) leverages a meta-learning framework to learn a regularization function that improves generalization to unseen domains. MLDG (Li et al. 2018) adapts meta-learning principles to train models that can rapidly generalize to new distributions. GroupDRO (Sagawa et al. 2020) minimizes the worst case loss among multiple source domains using the DRO framework. Unlike our implementation, it does not incorporate concept shift into the uncertainty set.

**4.3.4. Results and Discussion.** Tables 3 and 4 summarize the predictive performance of the ERM and different DG methods under two real-world settings: customer churn prediction and patient mortality prediction. Overall, all DG methods show modest improvements over the ERM baseline, suggesting that incorporating domain-level regularization or data augmentation can partially mitigate distributional discrepancies. However, these gains remain limited, reflecting the fact that real-world shifts are more complex than those typically simulated in benchmark data sets, which primarily capture covariate shifts only (Zhang et al. 2023b). Comparing across the two data sets, we observe higher performance lifts in the in-hospital mortality prediction task than in the customer churn prediction. This difference is consistent with the nature of their underlying shifts. The clinical data primarily exhibit covariate shifts (Zhang et al. 2021a); for instance, differences in patient demographics. In contrast, the churn data set involves temporal and behavioral shifts caused by marketing campaign rollouts and evolving customer engagement patterns, which introduce stronger concept shifts and, thus, pose a greater challenge for generalization.

Our illustrative DRO implementation, which explicitly incorporates both covariate and concept shift components in defining the uncertainty set, achieves improvements across both predictive tasks. The result demonstrates that optimizing for worst case distributions can enhance robustness without changing the underlying model architecture. Importantly, as with

**Table 3.** Customer Churn Prediction Performance

| | Serving time frame | | | | | | |
|---|---|---|---|---|---|---|---|
| Method | 2013-09 | 2013-10 | 2013-11 | 2013-12 | 2014-01 | 2014-02 | Average (lift, %) |
| ERM (Vapnik 1991) | 0.733 | 0.631 | 0.668 | 0.617 | 0.639 | 0.666 | 0.659 |
| ADA (Volpi et al. 2018) | 0.744 | 0.650 | 0.668 | 0.639 | 0.645 | 0.658 | 0.667 (1.21) |
| Mixup (Zhang et al. 2018) | 0.738 | 0.652 | 0.650 | 0.622 | 0.634 | 0.666 | 0.660 (0.15) |
| IRM (Arjovsky et al. 2019) | 0.744 | 0.629 | 0.661 | 0.634 | 0.633 | 0.670 | 0.662 (0.46) |
| MetaReg (Balaji et al. 2018) | 0.735 | 0.626 | 0.678 | 0.638 | 0.636 | 0.667 | 0.663 (0.61) |
| MLDG (Li et al. 2018) | 0.734 | 0.656 | 0.678 | 0.633 | 0.636 | 0.678 | 0.669 (1.52) |
| GroupDRO (Sagawa et al. 2020) | 0.732 | 0.648 | 0.675 | 0.642 | 0.642 | 0.679 | 0.670 (1.67) |
| DRO (illustrative implementation) | 0.769*** | 0.664*** | 0.692*** | 0.650*** | 0.664*** | 0.697*** | 0.689*** (4.55) |

*Notes.* Performance is measured by AUROC. The "Average" column reports the mean performance across serving time frames, whereas "lift" denotes the relative improvement over the ERM baseline. Statistical significance is assessed using a $t$-test against ERM.
***$p < 0.001$.

**Table 4.** In-Hospital Mortality Prediction Performance

| | Serving region | | | | |
|---|---|---|---|---|---|
| Method | West | South | Northeast | Midwest | Average (lift, %) |
| ERM (Vapnik 1991) | 0.864 | 0.876 | 0.819 | 0.850 | 0.852 |
| ADA (Volpi et al. 2018) | 0.902 | 0.918 | 0.837 | 0.876 | 0.883 (3.64) |
| Mixup (Zhang et al. 2018) | 0.891 | 0.896 | 0.867 | 0.869 | 0.881 (3.40) |
| IRM (Arjovsky et al. 2019) | 0.885 | 0.909 | 0.844 | 0.885 | 0.881 (3.40) |
| MetaReg (Balaji et al. 2018) | 0.904 | 0.916 | 0.851 | 0.899 | 0.893 (4.81) |
| MLDG (Li et al. 2018) | 0.870 | 0.917 | 0.823 | 0.869 | 0.870 (2.11) |
| GroupDRO (Sagawa et al. 2020) | 0.878 | 0.903 | 0.829 | 0.863 | 0.868 (1.88) |
| DRO (illustrative implementation) | 0.923*** | 0.934*** | 0.890*** | 0.908*** | 0.914*** (7.28) |

*Notes.* Performance is measured by AUROC. Each column reports performance when the region specified by that column is treated as the serving domain with the remaining regions used as source training domains. The “Average” column reports the mean performance across serving regions, whereas “lift” denotes the relative improvement over the ERM baseline. Statistical significance is assessed using a *t*-test against ERM.

***$p < 0.001$.

other DG methods, the DRO implementation also uses standard cross-validation for hyperparameter tuning. By explicitly optimizing against potential adverse scenarios, DRO mitigates the risk of overfitting to the empirical training distribution.

Nevertheless, it is important to note that, when the actual distributional shift at deployment exceeds the range inferred from the training domains, the model may still experience performance degradation, similar to ERM. Therefore, it remains a challenge and a promising research direction to develop data-driven uncertainty estimation and adaptive DRO formulations that can dynamically expand or reshape the uncertainty set as new data accumulates during deployment.

## 5. Practical Recommendations for Robust Predictive Modeling

We summarize four practical recommendations to help IS researchers design, evaluate, and apply predictive models more robustly in the presence of uncertainty.

i. First principles thinking for predictive modeling: design for anticipated shifts. When developing predictive models, researchers should be mindful of the types of distributional shifts most likely to occur in their focal task as it relates to their claimed knowledge contributions and incorporate corresponding inductive biases accordingly. If the task is expected to primarily experience covariate shift, the model should emphasize learning domain-invariant representations. For instance, in a hospital readmission prediction task, a model trained in one hospital may later be deployed in another region. Whereas the underlying clinical mechanisms linking patient attributes to readmission risk may remain stable, the demographic composition of patients can differ substantially. In such cases, learning representations that are invariant across hospitals helps prevent spurious correlations, such as overfitting to regional or demographic indicators that are not causally related to readmission outcomes. Conversely, if the prediction task is more likely to experience concept shift, model design should prioritize robustness against overfitting to the training label-generation mechanism. Regularization, smoothing-based methods, or simpler model architectures can mitigate overfitting to highly specific decision boundaries that may not generalize when the underlying relationship between inputs and outputs changes. Accordingly, researchers mindful of mitigating the effect of unseen distributional shifts can follow a three-step process of assessing, analyzing, and evaluating. This is especially crucial with the shift toward predicting microlevel outcomes and the greater impetus on responsible artificial intelligence (AI) (Brown et al. 2015, Krishnan et al. 2025).

ii. Assess potential serving distribution shifts. Before designing a novel predictive model or applying existing methods, researchers should first assess whether their focal predictive task is likely to experience unseen distributional shifts between the training and serving stages. For researchers who design predictive models, identifying potential sources of shifts helps determine whether robustness should be prioritized in model design. For instance, customer churn prediction or demand forecasting tasks or financial market prediction are inherently time sensitive. Changes in marketing strategies, user demographics, or external shocks can easily alter data distributions. In such cases, researchers should explicitly anticipate these dynamics and guide the entire model design cycle to account for potential uncertainty.

For researchers who apply predictive models for constructing variables that inform explanatory/causal theory development, assessing the potential for unseen shifts is equally critical. A model pretrained on data from a different period or platform may encode spurious relationships that no longer hold in the researcher’s focal data set. For example, a sentiment classification model trained on user posts collected in 2022 from an online knowledge-sharing platform may fail to generalize to more recent posts as the rise of generative AI has subtly altered writing styles and content expression patterns (Shan and Qiu 2025). Applying such a predictive model without reassessment could introduce measurement error into downstream econometric analysis, compromising construct validity.

Therefore, as the first step, researchers should explicitly assess the relevance of this issue for their research setting. They could compare basic distributional statistics, evaluate out-of-sample performance on a small validation set, or use quantitative measures such as the maximum mean discrepancy to detect potential data drift. This step is conceptually similar to verifying covariate balance in propensity score matching: before applying a model to new data, researchers should ensure that the training and serving samples data are comparable.

iii. Apply DG methods effectively. Once the unseen distribution shift problem is found to be relevant to the focal research setting, given the diverse range of DG techniques reviewed in Section 3, researchers may wonder which class of methods is most suitable for them. First, the choice depends on the nature of the data shift. When potential shifts primarily reflect input distributional changes (covariate shift), domain-invariant learning methods such as IRM are appropriate as they encourage invariant representations across domains. When shifts may also involve changes in the label-generation mechanism (concept shift), learning-based approaches, particularly uncertainty-aware optimization methods such as DRO, are more suitable. In contrast, data-based approaches (e.g., Mixup) offer simplicity and can be readily adopted to enrich training diversity, especially when training data are scarce, but their effectiveness against complex serving distribution shifts remains limited.

Second, the choice also depends on data availability, particularly domain granularity. Most DG methods, including our DRO-based implementation, rely on multiple source domains to infer cross-domain invariances. When natural domain groupings exist (e.g., regions, time periods, organizations), domain-invariant or learning-based methods can effectively leverage such structure. However, when only limited or single-domain data are available, data-based augmentation methods may be more practical, or clustering-based techniques can be employed to manually construct subdomains.

iv. Evaluate under uncertainty. Because the serving distribution is unknown, researchers can simulate how their DG-enriched predictive models might perform, relative to standard or alternative DG models, based on how the unseen environment might realistically change. For covariate shift, one can construct evaluation data sets by intentionally modifying the composition of features as well as properties of specific features. For example, resampling certain subpopulations (e.g., regions, customer segments, or time periods) can mimic adoption in a new market, whereas adding structured noise or masking specific features can reflect real-world issues such as missing data or new data collection procedures. Similarly, simulation may be conducted for concept shift by redefining the label-generation rule to reflect potential decision boundary changes. Qualitative diagnostics, such as t-SNE visualizations of feature representations or shifts in SHAP value patterns (as shown in Section 2), can help to assess the simulated shifts.

To make these stress scenarios systematic rather than ad hoc, researchers may also consider quantifying and reporting the degree of shift introduced in the simulated evaluation data sets. One useful metric is the proxy $\mathcal{A}$-distance (Glorot et al. 2011), which quantifies how easily a classifier can distinguish between two domains. Specifically, given two data sets, a simple classifier, such as a $k$-nearest neighbor or logistic regression model, is trained to discriminate between samples from the two domains, and its generalization error $\epsilon$ is used to compute the proxy distance as $d_{\mathcal{A}} = 2(1 - 2\epsilon)$. The intuition is straightforward: if the two data sets come from similar distributions, it should be difficult to distinguish them, resulting in an error rate close to 0.5 and, thus, a small $d_{\mathcal{A}}$. Conversely, if the domains are highly dissimilar, the classifier easily separates them, yielding a small $\epsilon$ and a $d_{\mathcal{A}}$ close to two.

In practice, researchers can use this metric to verify that the constructed simulation data sets indeed represent meaningful shifts of varying intensity. By gradually increasing the degree of simulated perturbation, such as progressively reweighting subpopulations or introducing stronger label definition changes, and measuring both model performance and the corresponding proxy $\mathcal{A}$-distance, one can plot a robustness curve that reveals tipping points at which predictive performance deteriorates sharply. In other words, an equally important question to average performance is how reliable a predictive model remains when things go wrong. Identifying such thresholds or robustness curves may provide valuable insights into the boundaries of a model's reliability under uncertainty.

## 6. Conclusion

Predictive modeling has become central to IS research, offering powerful tools for understanding and forecasting organizational and individual behaviors. Yet the implicit assumption that data are stable and independent and identically distributed often fails to hold in dynamic real-world environments. This commentary argues that distributional shift is not an exception but rather a defining characteristic of modern predictive settings. We call for a rethinking of how predictive models employed in IS research are designed, evaluated, and reported, moving from static evaluation toward validity-oriented design and robustness-centric machine learning variable construction.

The distributionally robust optimization framework exemplifies this shift in mindset. Rather than optimizing solely for average accuracy, DRO explicitly accounts for worst case scenarios within plausible distributional neighborhoods. Whereas not the only solution, it demonstrates how robustness thinking can be operationalized in predictive models employed for research to safeguard performance under uncertainty. We present several practical recommendations to guide the use, design, and evaluation of robust predictive models. By anticipating potential shifts throughout the theorized model life cycle, assessing data representativeness, stress-testing models under simulated uncertainty, and reporting worst case performance alongside averages, predictive models employed in modern research can become not only more credible but also more resilient to the realities of a changing world.

## Acknowledgments

The authors gratefully thank the senior editor, the associate editor, and anonymous reviewers for their constructive advice and guidance.

## Endnotes

[1] In the transfer learning literature, a domain refers to a specific data set or environment from which a data distribution is drawn.

[2] We thank the authors of Kitchens et al. (2018) for providing the data set.